\documentclass[10pt,twocolumn,letterpaper]{article}

\usepackage[pagenumbers]{cvpr}

\usepackage{graphicx}
\usepackage{amsmath}
\usepackage{cuted}

\usepackage{amssymb}
\usepackage{tikz}
\usepackage{pgfplots}
\usepackage{booktabs}
\usepackage{amsfonts}
\usepackage{bbding}
\usepackage{color}
\usepackage{multirow}
\usepackage{multicol}
\usepackage[accsupp]{axessibility}
\usepackage{makecell}
\usepackage{colortbl}
\usepackage{caption}
\definecolor{Gray}{gray}{0.92}

\usepackage{graphics}
\usepackage{epsfig}
\usepackage{mathptmx}
\usepackage{times}
\usepackage{amsmath}
\usepackage{amssymb}
\usepackage{siunitx}
\usepackage{newtxtext,newtxmath}

\usepackage{times}
\usepackage{helvet}
\usepackage{courier}
\usepackage[hyphens]{url}
\usepackage{graphicx}
\usepackage{caption}
\usepackage{amsmath}
\usepackage{amssymb}
\usepackage{colortbl}
\usepackage{subcaption}
\usepackage{booktabs} 
\usepackage{multirow}
\usepackage[dvipsnames, svgnames, x11names]{xcolor}
\definecolor{lgray}{gray}{0.7}
\usepackage{algorithm}
\usepackage{algorithmic}
\usepackage{enumitem}
\usepackage{bibentry}

\usepackage{newfloat}
\usepackage{listings}

\definecolor{cvprblue}{rgb}{0.21,0.49,0.74}

\usepackage[pagebackref,breaklinks,colorlinks]{hyperref}

\usepackage[capitalize]{cleveref}
\crefname{section}{Sec.}{Secs.}
\Crefname{section}{Section}{Sections}
\Crefname{table}{Table}{Tables}
\crefname{table}{Tab.}{Tabs.}

\def\confName{CVPR}
\def\confYear{2024}

\DeclareUnicodeCharacter{0301}{\'{e}}

\begin{document}

\title{Video4Edit: Viewing Image Editing as a Degenerate Temporal Process}
\vspace{-2cm}
\author{
Xiaofan Li \quad Yanpeng Sun \quad Chenming Wu \quad YuAn Wang \\
Fan Duan \quad Weihao Bo \quad Yumeng Zhang \quad Dingkang Liang \\
\textsuperscript{} \textbf{Baidu Inc.}\\
}

\maketitle

\begin{strip}
\begin{minipage}{\textwidth}\centering
\vspace{-30pt}
\includegraphics[width=0.94\textwidth]{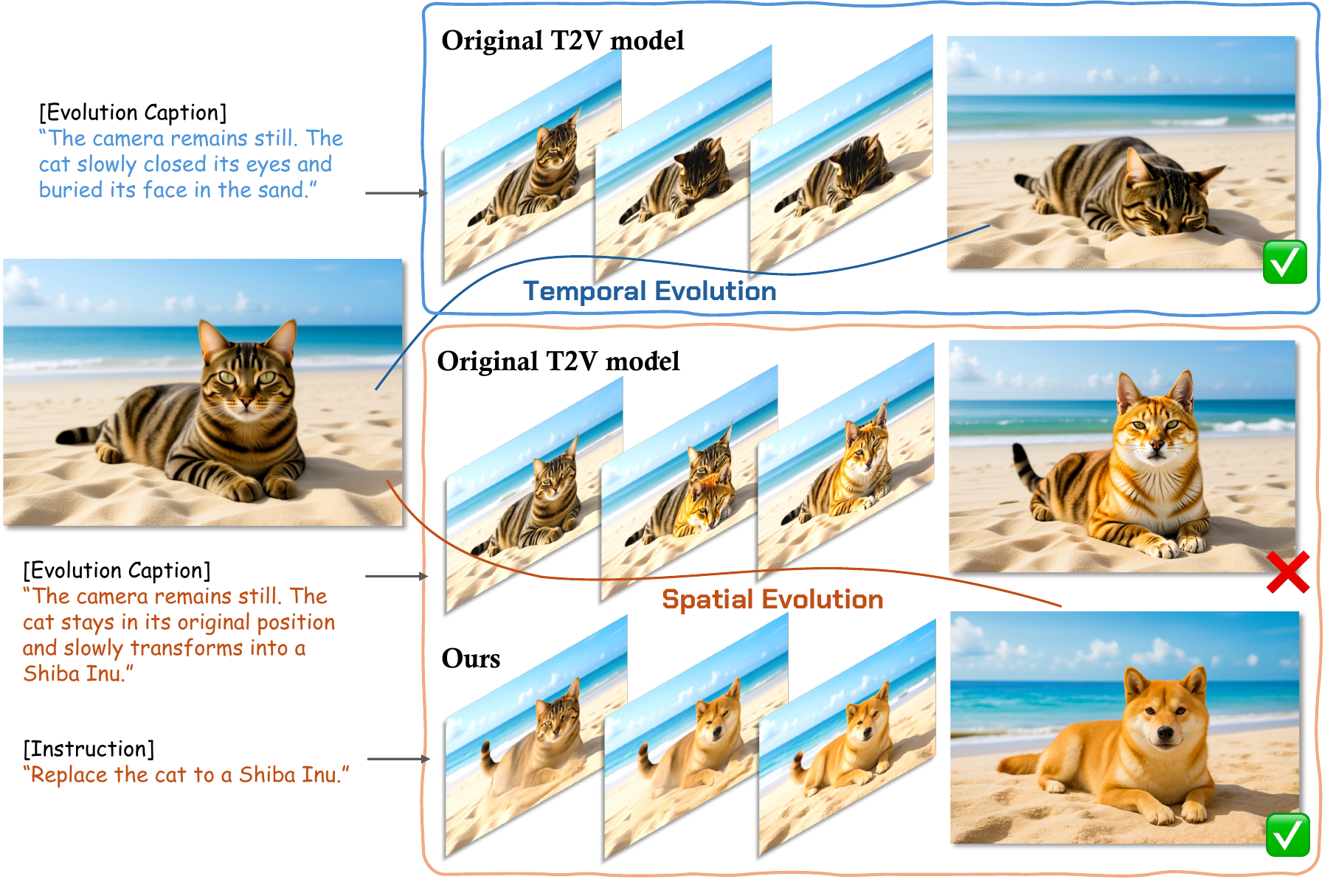}
\captionof{figure}{\textbf{Video4Edit: image editing as a degenerate temporal process.} We view image edits through a temporal lens and categorize them into two families: \emph{temporal evolution} (state changes over time with minimal spatial re-layout) and \emph{spatial evolution} (structural reconfiguration). After rewriting the instruction into an evolution-style caption, a video-pretrained T2V model can often perform temporal-evolution edits in a zero-shot manner (though tasks such as replace still need additional consistency constraints), while spatial-evolution edits remain challenging. We find that a light fine-tuning of the video-pretrained model suffices to handle both families, enabling general-purpose image editing.}
\label{fig:teaser}
\end{minipage}
\end{strip}
\begin{abstract}
We observe that recent advances in multimodal foundation models have propelled instruction-driven image generation and editing into a genuinely cross-modal, cooperative regime. Nevertheless, state-of-the-art editing pipelines remain costly: beyond training large diffusion/flow models, they require curating massive high-quality triplets of \{instruction, source image, edited image\} to cover diverse user intents. Moreover, the fidelity of visual replacements hinges on how precisely the instruction references the target semantics. We revisit this challenge through the lens of temporal modeling: if video can be regarded as a full temporal process, then image editing can be seen as a degenerate temporal process. This perspective allows us to transfer single-frame evolution priors from video pre-training, enabling a highly data-efficient fine-tuning regime. Empirically, our approach matches the performance of leading open-source baselines while using only about one percent of the supervision demanded by mainstream editing models.
\end{abstract}

\section{Introduction}
Instruction-driven image editing has made rapid progress with diffusion/flow models, but state-of-the-art pipelines are still data-hungry: training commonly relies on massive triplets of \{instruction, source image, edited image\} curated by synthetic or human-in-the-loop procedures \citep{brooks2023instructpix2pix,zhang2023magicbrush,hui2024hqedit,zhao2024ultraedit,yu2024anyedit,ge2024seeddataedit,wang2025gptimageedit}. At the same time, unified multimodal models (UMMs) that couple understanding and generation promise a single architecture for both text and pixels \citep{xiao2024omnigen,batifol2025kontext,shi2025seededit3,wang2025lightbagel,liu2025step1x}, yet their editing accuracy in the low-data regime typically lags behind specialized editors trained with large triplets. Training-free or inversion-based alternatives---e.g., Prompt-to-Prompt and attention-control variants \citep{hertz2022prompt}, SDEdit \citep{meng2021sdedit}, Imagic \citep{kawar2023imagic}, Null-text inversion \citep{mokady2023null}, external control via ControlNet \citep{Zhang2023AddingCC}, and recent flow-matching editors \citep{bar2025flowedit}---alleviate data collection but still struggle to jointly preserve structure and non-edit regions when edits are complex. Meanwhile, the foundation of high-fidelity T2I generation continues to evolve from DDPMs to Latent Diffusion, DiT, SDXL, and SD3 \citep{ho2020denoising,Rombach2021HighResolutionIS,peebles2023scalable,podell2024sdxl,esser2024scaling}, providing stronger generative backbones yet not fully resolving edit-locality and context consistency. Training high-quality instruction-driven image editors typically hinges on the same large triplet corpora noted above, and while recent UMMs reduce engineering overhead, their editing accuracy in practice still trails specialized editors such as Step1X-Edit and LightBagel. This raises a central question: can we close the gap without scaling triplets data to millions?

Motivated by these limitations, we revisit editing through a temporal lens: if video generation is a full temporal process, then image editing can be viewed as a short, degenerate evolution from the source to the target. Empirically (Fig.~\ref{fig:teaser}), a video-pretrained T2V model can execute many temporal-evolution edits in a zero-shot manner once the instruction is rewritten as an evolution-style caption.
However, experiments proved that two capabilities in video generators are insufficient for editing: (i) spatial evolution, i.e., object- or layout-level changes that require geometric consistency; and (ii) non-edit-region consistency, i.e., keeping context identical to the source outside the edit locus. Addressing both requires temporally grounded guidance and explicit anchoring of the final outcome.

We propose Video4Edit, a data-efficient teacher–student pipeline built on video-pretrained backbones. A frozen, video-pretrained FLF2V teacher receives the source/edited pair as the first/last frames, together with an evolution caption, to roll out temporally coherent intermediate states; a video-pretrained, trainable I2V student takes only the source image and instruction. Training combines two complementary objectives: (1) block-wise L2 distillation on DiT hidden states to transfer temporal priors; and (2) tail-frame supervision computed after flow-domain inversion and 3D‑VAE encoding, with timestep weighting to emphasize late stages. At inference, only the student is used; no teacher frames or evolution captions are needed.
Despite using roughly 1\% of mainstream supervision---still far smaller than recent large-scale corpora \citep{zhao2024ultraedit,ge2024seeddataedit,wang2025gptimageedit}---Video4Edit delivers competitive performance on both GEdit-Bench-EN and ImgEdit-Bench \citep{liu2025step1x,ye2025imgedit}, narrowing the gap to specialized editors while retaining the flexibility of an instruction-driven, mask-free pipeline.
In summary, we make three primary contributions:
\begin{itemize}
\item We innovatively regard instruction-driven image editing as a degenerate temporal process, showing that video priors can often support zero-shot temporal-evolution edits while still exposing gaps in spatial evolution and context preservation.
 \item We introduce Video4Edit, a teacher–student framework that couples block-wise DiT distillation with flow-inverted tail supervision in 3D‑VAE latent space, distilling the teacher's evolution trajectory into the student while explicitly enforcing edit locality and non-edit-region consistency.
  \item With roughly 1\% of mainstream supervision, Video4Edit matches or surpasses strong open-source baselines on GEdit-Bench-EN and achieves first/second rank across most ImgEdit-Bench categories.
\end{itemize}

\begin{figure*}[t]
\centering
\includegraphics[width=\textwidth]{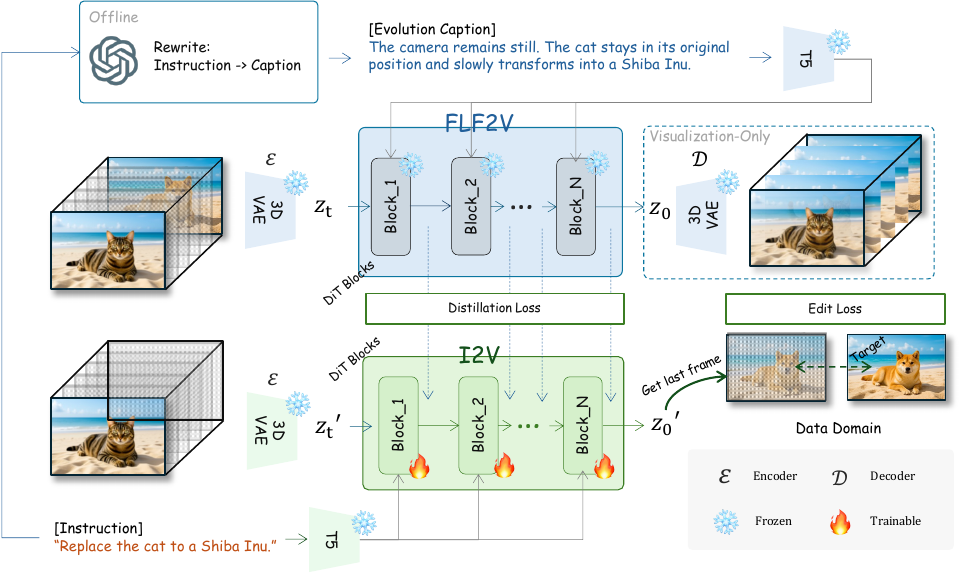}
\caption{\textbf{Video4Edit overall pipeline.} We formulate image editing as a degenerate temporal process and adopt a teacher–student framework. The teacher (Wan2.1 FLF2V-14B~\citep{wan2025}) receives the source image as the first frame and the edited image as the last frame, guided by an offline evolution prompt distilled from the instruction, to roll out temporally coherent intermediate states. The student (Wan2.1 I2V-14B-720P) takes only the source image and instruction, learning from teacher signals to produce the edited result in a few steps at inference.}
\label{fig:m1}
\end{figure*}

\section{Related Work}

\subsection{Instruction-driven Diffusion/Flow Editing.}
Early instruction-driven editing systems typically relied on loosely coupled "language-to-diffusion" pipelines: methods such as InstructPix2Pix \citep{brooks2023instructpix2pix} and MagicBrush \citep{zhang2023magicbrush} first synthesized paired data via large language models or human annotation, and then invoked diffusion models for local resynthesis. Recent works, including SmartEdit, AnyEdit, and UltraEdit \citep{zhao2024ultraedit}, tighten the semantic–visual coupling through task-aware routing and fine-grained attention control. In parallel, training-free techniques like Null-text inversion \citep{mokady2023null} and structure-preserving constraints \citep{huberman2024edit} aim to improve editability on real images without retraining, while FlowEdit \citep{bar2025flowedit} proposes an inversion-free path for flow models, highlighting the remaining challenge of jointly preserving structure and textual fidelity on large-scale models such as SD3 and FLUX.

\subsection{Data-centric Editing Corpora.}
As instruction complexity and multi-turn interactions increase, dataset construction has evolved from purely manual labeling toward a hybrid synthesis-and-real pipeline. MagicBrush \citep{zhang2023magicbrush} ensures accuracy through human annotation, whereas InstructPix2Pix \citep{brooks2023instructpix2pix} and HIVE leverage LLMs and Prompt-to-Prompt \citep{hertz2022prompt} to scale synthetic pairs. HQ-Edit \citep{hui2024hqedit} and UltraEdit \citep{zhao2024ultraedit} further employ GPT-4V and DALL·E 3 to improve visual fidelity and instruction alignment. Building on this, \textsc{GPT-Image-Edit-1.5M} \citep{wang2025gptimageedit} consolidates OmniEdit/HQ-Edit/UltraEdit and rewrites prompts, while SEED-Data-Edit \citep{ge2024seeddataedit} blends automated pipelines, real user requests, and expert multi-turn edits to better match in-the-wild distributions. SeedEdit3.0 \citep{shi2025seededit3} introduces a "meta-tag + recaption + task-label" paradigm that balances synthetic diversity with real-case robustness, providing reusable practices for data preparation in unified editing systems.

\subsection{Multimodal Editing Architectures.}
Architecturally, the community is moving from parallel autoregressive–diffusion exploration toward unified multimodal frameworks. Autoregressive models (e.g., ControlAR, ControlVAR) excel at localized constraints with discrete tokens but suffer from sequence-length bottlenecks; diffusion models, after the DDPM → Latent Diffusion → DiT evolution, dominate high-fidelity generation yet still lack robust multi-turn alignment. Recent advances integrate understanding, generation, and editing within a single pipeline: SeedEdit3.0 \citep{shi2025seededit3} employs meta-information tagging and reward modeling to fuse heterogeneous data; Step1X-Edit \citep{liu2025step1x} places an MLLM front-end for semantic analysis and introduces GEdit-Bench for user-centric evaluation; FLUX.1 Kontext \citep{batifol2025kontext} and LightBagel \citep{wang2025lightbagel} realize unified generation/editing via flow matching and dual-branch fusion, respectively. Despite improved consistency and responsiveness, these systems still incur substantial data and training costs, motivating research on low-data, highly generalizable alternatives.

\begin{table*}[t]
\centering
\small
\setlength{\tabcolsep}{3.5pt}
\caption{Mapping common editing tasks to evolution-style caption templates for the teacher. The templates turn a concise instruction into a short, temporally grounded description that guides intermediate states.}
\label{tab:evolution_prompt_mapping}
\resizebox{\textwidth}{!}{
\begin{tabular}{p{3.1cm} p{5.8cm} p{8.4cm}}
\toprule
\textbf{Editing task} & \textbf{Typical instruction cues} & \textbf{Evolution-style caption template (for teacher)} \\
\midrule
Subject Addition & ``add a red balloon above the boy'' & The red balloon gradually appears above the boy, while everything else remains still. \\
Subject Removal & ``remove the trash can on the left'' & The trash can on the left gradually fades away, while everything else remains still. \\
Subject Replacement & ``replace the mug with a teacup'' & The mug gradually turns into a teacup in place, while pose, lighting, and surroundings remain unchanged. \\
Background Change & ``change background to a beach'' & The background slowly transforms into a beach scene, while the main subject and foreground remain unchanged. \\
Color Alteration & ``make the car blue'' & The car's body color gradually shifts to blue, while reflections and all other elements remain consistent. \\
Material Modification & ``turn the sofa into leather'' & The sofa's texture gradually becomes leather, while geometry and the rest of the scene remain unchanged. \\
Text Modification & ``change text to `OPEN' '' & The sign's text gradually changes to ``OPEN'', while layout and surrounding pixels remain unchanged. \\
Motion Change & ``raise the person's right hand'' & The person's right forearm slowly raises, while the rest of the body and scene remain steady. \\
Portrait Editing \& Beautification & ``smooth skin and brighten eyes'' & Facial skin is gently smoothed and eyes brightened over time, while identity and other details remain unchanged. \\
Style Transfer & ``convert to watercolor style'' & The scene gradually adopts a watercolor style, while composition and content remain unchanged. \\
Tone Transformation & ``apply warm cinematic grade'' & The image gradually shifts to a warm cinematic grade, while structure and content remain unchanged. \\
\bottomrule
\end{tabular}}
\end{table*}

\section{Video4Edit}
\subsection{Overall Pipeline}
We cast image editing as a degenerate temporal process: given a source image and a target edit, the transformation can be modeled as a short video evolving from the initial to the final state. Our pipeline adopts a teacher–student design that leverages a strong video generator as a temporal teacher to provide the evolution process from the source image to the edited image to supervise a student for data‑efficient learning, as illustrated in Fig.~\ref{fig:m1}. Both teacher and student are instantiated from the Wan2.1 family of video generative models. At inference, only the student is used. Given a source image and an instruction, the student rolls out a short trajectory and outputs the edited result in a few steps.

\paragraph{FLF2V as Teacher.}
We use Wan2.1 FLF2V‑14B~\citep{wan21github,wan21flf2v} as the teacher model. Here, ``FLF2V-14B'' denotes a frozen first/last-frame-conditioned video generator with 14B parameters. For each training triplet \{instruction, source image, edited image\}, we feed the source image as the first frame and the edited image as the last frame, respectively, and condition the model with an offline-constructed evolution prompt that describes a plausible step-by-step transformation consistent with the instruction. The teacher rolls out intermediate frames (and/or latent velocities) that define a temporally coherent editing trajectory. 

\vspace{-5mm}

\paragraph{I2V as Student.}
We use Wan2.1 I2V‑14B‑720P~\citep{wan21i2v} as the student model. Here, ``I2V-14B-720P'' denotes a trainable image-to-video model with 14B parameters and 720p resolution. The student only receives the source image as the first frame and the raw instruction as text input. During training, the teacher remains frozen and only the student parameters are updated, learning to follow the teacher's temporal signals while producing the final edited outcome.

\subsection{Evolution Prompt}
To bridge the gap between concise still-image instructions and the step-wise guidance expected by a temporal model, we run an Instruction-to-Caption Generator (ICG) offline. The ICG normalizes entity references, identifies the task type, decomposes the instruction into causally ordered atomic edits (e.g., color change, background replacement, geometry adjustment), and verbalizes a short evolution caption that emphasizes the transition rather than only the end state. For example, ``Remove the object'' becomes ``The object gradually fades away while everything else remains still.'' and ``Change the background to a forest'' becomes ``The background slowly transforms into a dense forest, with all other elements unchanged.'' These captions condition the teacher to produce semantically aligned intermediate frames, which in turn provide rich supervision for the student; the full library of task-to-caption mappings is documented in Tab.~\ref{tab:evolution_prompt_mapping}.

\paragraph{Prompting templates for ICG.}
We adopt a concise two-step prompting scheme aligned with Supplementary Table~S1:
\begin{enumerate}
  \item \textbf{Instruction $\rightarrow$ Evolution Caption.} Convert the instruction into a descriptive prompt suitable for video generation; emphasize the \emph{temporal process} rather than only the final state.\\
  Examples: ``Remove the object'' $\rightarrow$ ``The object gradually fades away while everything else remains still.''; ``Change the background to a forest'' $\rightarrow$ ``The background slowly transforms into a dense forest, with all other elements unchanged.'' See Tab.~\ref{tab:evolution_prompt_mapping} for additional templates.
  \item \textbf{Post‑refinement.} Lightly polish the caption (without altering semantics) to stress \emph{high resolution}, \emph{high quality}, and \emph{globally coherent temporal evolution}.\\
  Template: ``\emph{OUTPUT}. The video is high‑resolution and high‑quality, and the overall evolution is temporally coherent.''
\end{enumerate}

\subsection{Data Coverage}
To demonstrate that our training corpus is both task-balanced and data-efficient, we report: (i) the per-task distribution across the 11 common editing categories together with the mix across the three public datasets we use (Tab.~\ref{tab:data_diversity}); and (ii) a scale comparison against representative prior works (Tab.~\ref{tab:external_scale}). Tab.~\ref{tab:data_diversity} details how OmniEdit, HQ-Edit, and UltraEdit are each sampled via balanced subsets, yielding nearly uniform coverage (standard deviation $\approx$1\%) over the 11 canonical edit types ranging from subject-level manipulations to tone transformation. Tab.~\ref{tab:external_scale} further positions our $<\!10$k supervision against large-scale corpora (e.g., UltraEdit $\sim$4M, LightBagel $\sim$45M), underscoring the $<\!1\%$ relative scale. We also summarize the open-source datasets adopted by representative works, together with their reported scales, and list the public datasets we use alongside our consumption policy. When a method does not disclose per-task or per-source breakdown, we keep the official aggregate size for reference.

\begin{table}[t]
\centering
\small
\setlength{\tabcolsep}{1.8pt}
\caption{Representative editing corpora, their public sources, and reported scale. Our supervision stays within one percent of mainstream datasets.}
\label{tab:external_scale}
\resizebox{\linewidth}{!}{
\begin{tabular}{l p{4.5cm} r c}
\toprule
\textbf{Work} & \textbf{Sources (examples)} & \textbf{Pairs / Triplets} & \textbf{Relative to Ours} \\
\midrule
FlowEdit & DIV2K + royalty-free photos (real images) & $\sim70$ imgs / $\sim250$ pairs & -- \\
UltraEdit & LLM-generated instructions + real images & $\sim4.0\,\mathrm{M}$ & $>\!400\times$ \\
SEED-Data-Edit & Auto pipelines + real-user requests + expert multi-turn & $\sim3.7\,\mathrm{M}$ & $>\!350\times$ \\
GPT-Image-Edit-1.5M & OmniEdit, HQ-Edit, UltraEdit (regenerated/rewritten) & $1.5\,\mathrm{M}$ & $>\!150\times$ \\
Step1X-Edit (HQ) & Aggregated multi-source editing corpus & $>\!1.0\,\mathrm{M}$ & $>\!100\times$ \\
LightBagel (mixed) & BLIP-3o, Civitai, OmniGen/OmniEdit, GPT-IE-1.5M, UniWorld-V1 & $\sim45\,\mathrm{M}$ corpus & $>\!4500\times$ \\
FLUX.1 Kontext & Internal (KontextBench released) & N/A & N/A \\
Qwen-Image & Multi-stage curated (text rendering + general) & N/A & N/A \\
\midrule
\textbf{Ours} & \textbf{Balanced subset of OmniEdit, HQ-Edit, UltraEdit} & \textbf{$<\!10$k} & \textbf{1$\times$ (baseline)} \\
\bottomrule
\end{tabular}}
\end{table}

\begin{table}[t]
\centering
\small
\setlength{\tabcolsep}{2.4pt}
\caption{Dataset mix (three public sources) and per-task distribution (11 categories) in our training split, highlighting diverse coverage with balanced supervision.}
\label{tab:data_diversity}
\resizebox{\linewidth}{!}{
\begin{tabular}{lcc}
\toprule
\textbf{Entry} & \textbf{Share / Usage} & \textbf{Notes} \\
\midrule
\multicolumn{3}{l}{\textit{Dataset mix (3 public sources)}} \\
OmniEdit & $\approx$34\% & instruction-style pairs; general edits \\
HQ-Edit & $\approx$33\% & high-fidelity, instruction-aligned pairs \\
UltraEdit & $\approx$33\% & large-scale, fine-grained edits \\
\midrule
\multicolumn{3}{l}{\textit{Task coverage (11 categories)}} \\
Subject Addition & 7.6\% & object appears gradually \\
Subject Removal & 9.8\% & object fades/inpainted \\
Subject Replacement & 9.3\% & semantic swap in place \\
Background Change & 11.1\% & re-layout background only \\
Color Alteration & 7.9\% & attribute shift \\
Material Modification & 12.2\% & texture/BRDF change \\
Text Modification & 7.2\% & glyph/perspective stable \\
Motion Change & 9.7\% & articulated pose change \\
Portrait Editing \& Beautification & 10.5\% & identity preserved \\
Style Transfer & 7.5\% & global style, content fixed \\
Tone Transformation & 8.3\% & grading/weather/light \\
\bottomrule
\end{tabular}}
\end{table}

\subsection{Training Objectives}
We optimize a compact objective combining two losses: (i) block‑wise hidden‑state distillation on the DiT backbone (L2 on selected blocks); and (ii) tail‑frame latent supervision computed after flow‑domain inversion and 3D‑VAE encoding, which drives edit‑region convergence and non‑edit‑region consistency with the source image. The total objective is:
\begin{equation}
\mathcal{L} \;=\; \lambda_{\text{kd}}\mathcal{L}_{\text{kd}} \;+\; \lambda_{\text{tail}}\mathcal{L}_{\text{tail}}.
\end{equation}
Here, \(\lambda_{\text{kd}}\) and \(\lambda_{\text{tail}}\) are loss weights; \(\mathcal{L}_{\text{kd}}\) and \(\mathcal{L}_{\text{tail}}\) denote the DiT block-wise distillation loss and the tail-frame supervision loss in 3D-VAE latent space, respectively.

\paragraph{Block‑wise distillation on DiT.}
For each selected transformer block \(b\) (and layer index \(\ell\)) in the DiT backbone, the student hidden states are directly aligned to the teacher via an L2 objective:
\begin{equation}
\mathcal{L}_{\text{kd}} \;=\; \sum_{b\in\mathcal{B}} \big\|h_{b}^{\text{stu}} - h_{b}^{\text{tea}}\big\|_2^2.
\end{equation}
We apply it on sparse blocks to reduce cost while stabilizing training through intermediate representation alignment.

\paragraph{Tail‑frame supervision in 3D‑VAE latent.}
The 3D‑VAE compresses every 4 frames into one latent block. To supervise the final outcome under flow‑matching training, we map the student's predictions at selected timesteps \(t\in\mathcal{T}_{\text{tail}}\) back to the data domain via flow inversion:
\begin{equation}
\begin{aligned}
\hat{x}_0 \;&=\; x_t \;-\; t\,\hat{u}(x_t,t) && \text{(Rectified Flow)}\\
\hat{x}_0 \;&=\; x_t \;-\; \frac{\rho(t)}{\rho'(t)}\,\hat{u}(x_t,t) && \text{(general CFM)}
\end{aligned}
\end{equation}
where \(\hat{u}\) is the predicted velocity, \(\rho(t)\) the schedule, and \(x_t\) the state at time \(t\). We then tile the ground‑truth edited image 4 times to form a 4‑frame clip, and encode with the 3D‑VAE:
\begin{equation}
\begin{aligned}
z^{\text{gt}}_{\text{tail}} \;&=\; \mathrm{3DVAE}\big(\underbrace{x_{\text{edit}},\ldots,x_{\text{edit}}}_{\times 4}\big),\\
z^{\text{stu}}_{\text{tail}}(t) \;&=\; \mathrm{3DVAE}\big(\underbrace{\hat{x}_0(t),\ldots,\hat{x}_0(t)}_{\times 4}\big).
\end{aligned}
\end{equation}
To emphasize late‑time supervision while keeping earlier steps informative, we apply a timestep weight
\begin{equation}
w(t) \;=\; \big(1 - t_{\text{norm}}\big)^{3}, \qquad t_{\text{norm}} \,=\, t / T,
\end{equation}
where \(T\) denotes the maximum training timestep used for normalization, and compute the weighted latent loss
\begin{equation}
\begin{aligned}
\mathcal{L}_{\text{tail}}
&= \sum_{t\in\mathcal{T}_{\text{tail}}} w(t)\,\big[
\|z^{\text{stu}}_{\text{tail}}(t)-z^{\text{gt}}_{\text{tail}}\|_1 \\
&\quad + \beta\,\|\nabla z^{\text{stu}}_{\text{tail}}(t)-\nabla z^{\text{gt}}_{\text{tail}}\|_1 \big].
\end{aligned}
\end{equation}
which encourages edit‑region convergence and non‑edit‑region consistency with the source image; the optional gradient term, scaled by \(\beta\), stabilizes fine‑scale structures.

\section{Experiments}
\subsection{Experimental Settings}
We use AdamW (lr 1e‑4, weight decay 0.01, cosine decay with 5\% warm‑up), batch size 1 with gradient accumulation to an effective batch size of 8, EMA 0.9999, and bfloat16 mixed precision. For efficiency, 3D‑VAE tail features are precomputed and stored offline. At inference, we generate the edited result in 8 steps by default. To fit memory, we enable PyTorch FSDP with full parameter/gradient/optimizer sharding (ZeRO‑3 style), activation checkpointing, gradient accumulation, and FlashAttention‑2; the frozen teacher is served with tensor parallelism and no backprop, and we optionally offload optimizer states to CPU on 40–80\,GB GPUs.

\subsection{Quantitative Results}
\label{sec:image_editing}
\begin{figure}[t]
  \centering
  \includegraphics[width=0.48\textwidth]{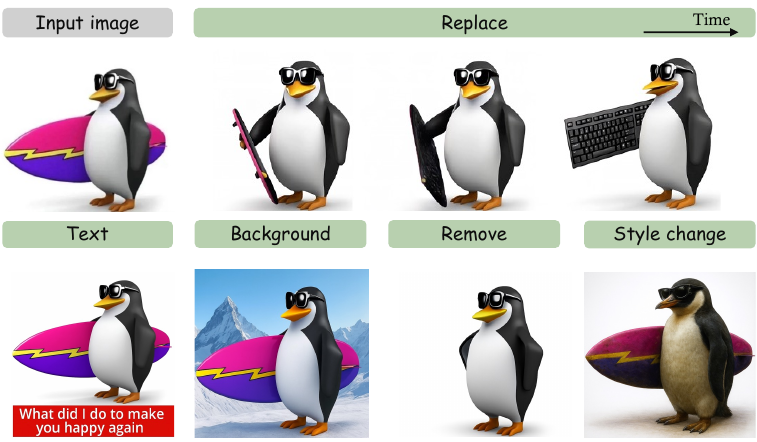}
  \caption{\textbf{Multi-task support.} Our method handles diverse editing tasks including subject addition, removal, replacement, background change, color alteration, and style transfer, demonstrating the versatility of our temporal-evolution framework.}
  \label{fig:multi_task}
\end{figure}

Following Step1X-Edit \citep{liu2025step1x} and the recent LightBagel \citep{wang2025lightbagel}, we evaluate on two editing suites that reflect complementary use cases: GEdit-Bench-EN \citep{liu2025step1x} focuses on in-the-wild user requests, while ImgEdit-Bench \citep{ye2025imgedit} decomposes editing into nine canonical task families (e.g., add, remove, replace). We compare against strong unified editors and recent specialized systems, treating LightBagel \citep{wang2025lightbagel} as a representative open-source baseline. Qualitative exemplars spanning these families are visualized in Fig.~\ref{fig:multi_task}, underscoring the breadth of tasks our method supports.

\paragraph{GEdit-Bench-EN.} Table \ref{tab:gedit} reports scores on real-user edits. Video4Edit delivers a high Overall (6.71) by coupling strong instruction alignment (SC 7.37) with solid perceptual quality (PQ 7.44). In the same setting, LightBagel reaches 6.06 Overall, confirming it as a strong but trailing baseline.

\paragraph{ImgEdit-Bench.} On the nine-category suite (Table \ref{tab:imgedit}), Video4Edit stays near the top across most task families—particularly Replace, Remove and Hybrid—leading to an Overall of 4.02. Notably, this advantage is achieved with roughly 1\% of the supervision commonly used by prior systems.

\begin{table}[t]
\centering
\footnotesize
\caption{Results on GEdit-Bench-EN (real-user edits). Higher is better ($\uparrow$). Best in \textbf{bold}, second best \underline{underlined}.}
\begin{tabular}{lccc}
\toprule
\textbf{Model} & \textbf{SC} $\uparrow$ & \textbf{PQ} $\uparrow$ & \textbf{Overall} $\uparrow$ \\
\midrule
Gemini-2.0-flash & 6.73 & 6.61 & 6.32 \\
GPT-4o           & \textbf{7.85} & \textbf{7.62} & \textbf{7.53} \\
\midrule
Instruct-Pix2Pix & 3.58 & 5.49 & 3.68 \\
MagicBrush      & 4.68 & 5.66 & 4.52 \\
AnyEdit         & 3.18 & 5.82 & 3.21 \\
ICEdit          & 5.11 & 6.85 & 4.84 \\
Step1X-Edit     & 7.09 & 6.76 & 6.70 \\
OmniGen2        & 7.16 & 6.77 & 6.41 \\
BAGEL           & 7.36 & 6.83 & 6.52 \\
Ovis-U1         & --   & --   & 6.42 \\
UniPic          & 6.72 & 6.18 & 5.83 \\
UniPic 2.0      & -- & -- & 7.10 \\
UniWorld-V1     & 4.93 & 7.43 & 4.85 \\
LightBagel      & 6.34 & 7.31 & 6.06 \\
\midrule
\rowcolor{gray!15}
\textbf{Video4Edit (ours)} & \underline{7.37} & \underline{7.44} & \underline{6.71} \\
\bottomrule
\end{tabular}
\label{tab:gedit}
\end{table}

\begin{table*}[t]
\centering
\caption{Results on ImgEdit-Bench (nine editing task families). Higher is better ($\uparrow$). Best in \textbf{bold}, second best \underline{underlined}.}
\resizebox{\linewidth}{!}{
\begin{tabular}{lccccccccc c}
\toprule
\textbf{Model} & \textbf{Add} $\uparrow$ & \textbf{Adjust} $\uparrow$ & \textbf{Extract} $\uparrow$ & \textbf{Replace} $\uparrow$ & \textbf{Remove} $\uparrow$ & \textbf{Background} $\uparrow$ & \textbf{Style} $\uparrow$ & \textbf{Hybrid} $\uparrow$ & \textbf{Action} $\uparrow$ & \textbf{Overall} $\uparrow$ \\
\midrule
GPT-4o & \textbf{4.61} & \textbf{4.33} & \textbf{2.90} & 4.35 & 3.66 & \textbf{4.57} & \underline{4.93} & \textbf{3.96} & \textbf{4.89} & \textbf{4.20} \\
\midrule
MagicBrush     & 2.84 & 1.58 & 1.51 & 1.97 & 1.58 & 1.75 & 2.38 & 1.62 & 1.22 & 1.90 \\
Instruct-Pix2Pix & 2.45 & 1.83 & 1.41 & 2.01 & 1.44 & 1.44 & 3.55 & 1.20 & 1.46 & 1.88 \\
AnyEdit        & 3.18 & 2.95 & 1.14 & 2.49 & 2.21 & 2.88 & 3.82 & 1.56 & 2.65 & 2.45 \\
UltraEdit      & 3.44 & 2.81 & 2.00 & 2.96 & 2.45 & 2.83 & 3.76 & 1.91 & 2.98 & 2.70 \\
Step1X-Edit    & 3.88 & 3.41 & 1.76 & 3.40 & 2.83 & 3.16 & 6.63 & 2.52 & 2.52 & 3.06 \\
ICEdit         & 3.58 & 3.39 & 1.73 & 3.15 & 2.93 & 3.08 & 3.84 & 2.04 & 3.68 & 3.05 \\
\midrule
OmniGen2     & 3.74 & 3.54 & 1.77 & 3.21 & 2.77 & 3.57 & 4.81 & 2.30 & 4.14 & 3.43 \\
BAGEL        & 3.56 & 3.31 & 1.88 & 2.62 & 2.88 & 3.44 & 4.49 & 2.38 & 4.17 & 3.20 \\
Ovis-U1      & 4.12 & 3.92 & \underline{2.36} & 4.09 & 3.57 & 4.22 & 4.69 & 3.23 & 3.61 & 3.98 \\
UniPic       & 3.66 & 3.51 & 2.06 & 4.31 & 2.77 & 3.77 & 4.76 & 2.56 & 4.04 & 3.49 \\
UniPic 2.0   & - & - & - & - & - & - & - & - & - & \underline{4.06} \\
UniWorld-V1    & 3.82 & 3.66 & 2.31 & 3.45 & 3.02 & 2.99 & 4.71 & 2.96 & 2.74 & 3.26 \\
LightBagel   & 4.21 & 3.23 & 1.83 & \underline{4.55} & \underline{3.80} & 4.15 & 4.66 & 3.93 & 3.60 & 3.77 \\
\midrule
\rowcolor{gray!15}
\textbf{Video4Edit (ours)} & \underline{4.35} & \underline{3.70} & 1.95 & \textbf{4.60} & \textbf{3.85} & \underline{4.30} & 4.85 & \underline{3.95} & 3.70 & 4.02 \\
\bottomrule
\end{tabular}
}
\label{tab:imgedit}
\end{table*}

\begin{figure*}[t]
  \centering
  \includegraphics[width=\textwidth]{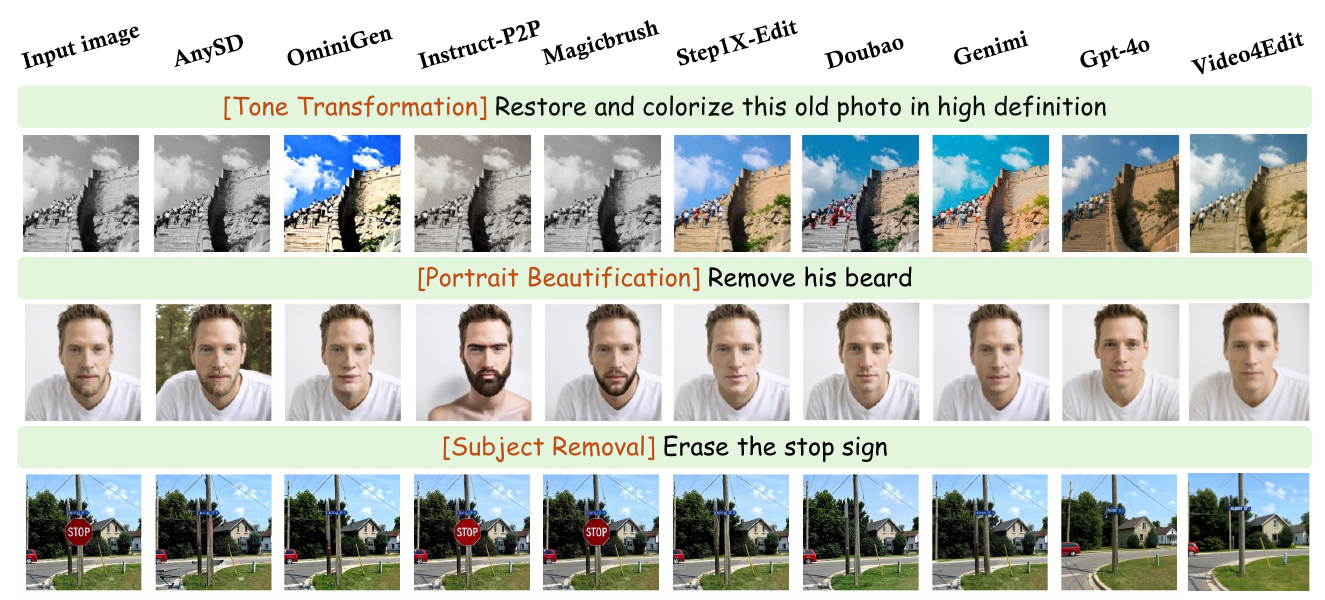}
  \caption{\textbf{A Comparative Illustration of Our Method, Open-Source Approaches, and Commercial Systems.}}
  \label{fig:qual_comp}
\end{figure*}

\subsection{Qualitative Results}
We present qualitative comparisons between our method and representative open‑source and commercial systems (Fig.~\ref{fig:qual_comp}). Open‑source baselines include OmniGen2~\citep{wu2025omnigen2}, BAGEL~\citep{deng2025emerging}, ICEdit~\citep{zhang2025ICEdit}, LightBagel~\citep{wang2025lightbagel}, and UniPic~\citep{wang2025skywork}; commercial systems include GPT‑4o and Gemini‑2.0‑flash. The visualization layout follows the presentation style of Step1X‑Edit~\citep{liu2025step1x} (all baselines are used in their publicly available configurations and are reproducible). Overall, our method better preserves non‑edit regions and delivers more faithful local edits under complex instructions, while maintaining strong perceptual quality.

\begin{figure*}[t]
  \centering
  \includegraphics[width=\textwidth]{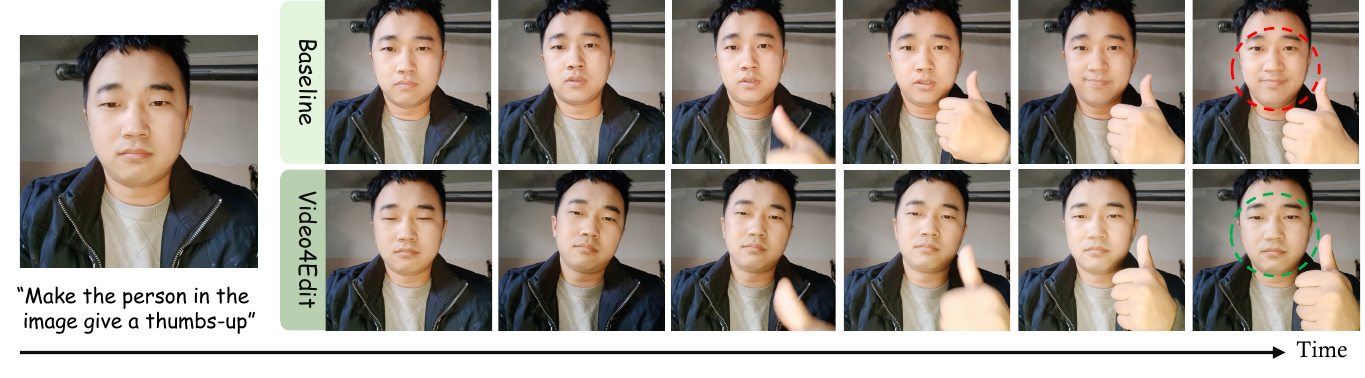}
  \caption{\textbf{Comparison with native I2V baseline.} Even in zero-shot scenarios where the native I2V model can generate plausible edits, it often introduces inconsistencies in non-edit regions (e.g., background artifacts, color shifts, structural distortions). Video4Edit maintains better consistency in non-edit regions through explicit supervision and distillation-based training.}
  \label{fig:i2v_baseline}
\end{figure*}

\subsection{Ablation Study}
\paragraph{Evaluation Protocol.}
We ablate supervision signals on GEdit-Bench-EN and ImgEdit-Bench, reporting Overall scores under each official protocol. Unless otherwise noted, we fix the backbone, training schedule, and inference steps, and vary a single factor at a time. The compared variants and their metrics are summarized in Tab.~\ref{tab:ablation_supervision}.

\paragraph{Effect of DiT Block-wise Distillation.}
Removing distillation from the frozen teacher causes the largest degradation (GEdit 6.71\,→\,6.05; ImgEdit 4.02\,→\,3.62). This indicates that transferring temporally grounded priors is crucial for stabilizing the student's short-horizon rollout and preserving edit locality. We also observe stronger structure preservation and fewer unintended changes in non-edit regions when distillation is enabled.

\vspace{-3mm}
\paragraph{Effect of Tail-frame Supervision.}
Discarding tail supervision also reduces performance (GEdit 6.71\,→\,6.28; ImgEdit 4.02\,→\,3.78). Without anchoring the final state via flow-domain inversion and 3D‑VAE encoding, the student exhibits late-stage drift and weaker identity/background consistency. Tail supervision complements block-wise distillation by explicitly constraining the outcome frame.

\vspace{-3mm}

\paragraph{Role of Evolution Captions (ICG).}
Feeding the teacher with raw instructions instead of evolution captions ("teacher w/o ICG") yields a smaller yet consistent drop (GEdit 6.71\,→\,6.45; ImgEdit 4.02\,→\,3.92). The instruction‑to‑caption generator provides temporal continuity cues, enabling the teacher to roll out smoother intermediate states and, consequently, cleaner distillation targets for the student.

\paragraph{Comparison with Native I2V Baseline.}
Even in scenarios where native I2V models can perform zero-shot editing, they often suffer from non-edit-region inconsistency issues. Fig.~\ref{fig:i2v_baseline} compares Video4Edit with the native I2V baseline (Wan2.1 I2V-14B-720P without fine-tuning) on zero-shot editing tasks. While the baseline can generate plausible edits, it frequently introduces unwanted changes in non-edit regions, such as background artifacts, color shifts, or structural distortions. Our method, through explicit tail-frame supervision and distillation-based training, maintains better consistency in non-edit regions while preserving edit quality, demonstrating the necessity of our fine-tuning approach even for zero-shot-capable scenarios.

\paragraph{Qualitative Observations.}
Without distillation, we often observe over‑editing and spurious changes in non‑edit regions; without tail supervision, outputs are visually plausible but deviate from the target semantics near the end of the rollout; without ICG, intermediate frames include small temporal jumps that translate into minor misalignments in the final image. These trends align with the quantitative results in Tab.~\ref{tab:ablation_supervision}.

\begin{table}[t]
\centering
\small
\setlength{\tabcolsep}{3pt}
\caption{Ablation on supervision components. Higher is better ($\uparrow$). Best in \textbf{bold}, second best \underline{underlined}. Metrics are Overall on GEdit-Bench-EN and ImgEdit-Bench.}
\label{tab:ablation_supervision}
\resizebox{\linewidth}{!}{
\begin{tabular}{lccc}
\toprule
\textbf{Variant} & \textbf{Distill} & \textbf{Tail sup.} & \textbf{GEdit / ImgEdit Overall} $\uparrow$ \\
\midrule
\rowcolor{gray!15}
Video4Edit (full) & $\checkmark$ & $\checkmark$ & \textbf{6.71 / 4.02} \\
\;\; w/o distillation &  & $\checkmark$ & 6.05 / 3.62 \\
\;\; w/o tail supervision & $\checkmark$ &  & 6.28 / 3.78 \\
\;\; teacher w/o ICG & $\checkmark$ & $\checkmark$ & \underline{6.45 / 3.92} \\
\bottomrule
\end{tabular}}
\end{table}

\section{Conclusion}

Video4Edit rethinks instruction-driven image editing as a degenerate temporal process and leverages a video-pretrained teacher--student pipeline to transfer temporal priors into a lightweight student editor. The combination of block-wise DiT distillation and tail-frame supervision enables faithful edits while using roughly $1\%$ of the supervision demanded by mainstream systems, delivering strong results across both in-the-wild and task-structured benchmarks.
Looking ahead, we plan to enrich the evolution-caption generator with multi-turn reasoning, extend the framework to video-to-video editing, and explore tighter integration with unified multimodal agents for interactive editing workflows. We believe that the visual temporal evolution patterns learned from large amounts of video provide excellent pre-training for visual understanding, editing, and reasoning.  
\vspace{-4mm}
\paragraph{Limitations.} Our model's parameter amount and flow matching inference requiring multiple steps limit deployment and inference speed. We explore acceleration strategies in the supplementary material; future work will focus on model compression and faster inference algorithms.

{
    \small
    \bibliographystyle{ieeenat_fullname}
    \bibliography{main}
}

\end{document}